\documentclass[conference]{IEEEtran}
\IEEEoverridecommandlockouts
\usepackage{graphics} 
\usepackage{epsfig} 
\usepackage{mathptmx} 
\usepackage{times} 
\usepackage{amsmath} 
\usepackage{amssymb}  
\usepackage{comment} 
\usepackage{hyperref}
\usepackage{color}
\usepackage{stfloats}
\usepackage{algorithmic}
\usepackage[linesnumbered,ruled,vlined]{algorithm2e}
\usepackage{mathtools}
\usepackage{multirow}
\DeclarePairedDelimiter{\ceil}{\lceil}{\rceil}

\begin{document}

\title{Non-Intrusive Load Monitoring for Feeder-Level EV Charging Detection: Sliding Window-based Approaches to Offline and Online Detection
\thanks{This work was supported in part by the Australian Research Council (ARC) Discovery Early Career Researcher Award (DECRA) under Grant DE230100046.}
}

\author{\IEEEauthorblockN{Cameron Martin\textsuperscript{1}, Fucai Ke\textsuperscript{2}, Hao Wang\textsuperscript{1,3*}}
\IEEEauthorblockA{
\textsuperscript{1}Department of Data Science and AI, Faculty of Information Technology, Monash University, Australia \\
\textsuperscript{2}Department of Human Centred Computing, Faculty of Information Technology, Monash University, Australia \\
\textsuperscript{3}Monash Energy Institute, Monash University, Australia
}
\thanks{*Corresponding author: Hao Wang (hao.wang2@monash.edu).}
}

\maketitle

\begin{abstract}
Understanding electric vehicle (EV) charging on the distribution network is key to effective EV charging management and aiding decarbonization across the energy and transport sectors. Advanced metering infrastructure has allowed distribution system operators and utility companies to collect high-resolution load data from their networks. These advancements enable the non-intrusive load monitoring (NILM) technique to detect EV charging using load measurement data. While existing studies primarily focused on NILM for EV charging detection in individual households, there is a research gap on EV charging detection at the feeder level, presenting unique challenges due to the combined load measurement from multiple households. In this paper, we develop a novel and effective approach for EV detection at the feeder level, involving sliding-window feature extraction and classical machine learning techniques, specifically models like XGBoost and Random Forest. Our developed method offers a lightweight and efficient solution, capable of quick training. Moreover, our developed method is versatile, supporting both offline and online EV charging detection. Our experimental results demonstrate high-accuracy EV charging detection at the feeder level, achieving an F-Score of 98.88\% in offline detection and 93.01\% in online detection. 
\end{abstract}


\section{Introduction} \label{intro}
With a fast-growing uptake of electric vehicles (EVs), the peak load on distribution networks is expected to increase significantly. A study in \cite{Muratori2018} revealed that, for a transformer connected to six households, an additional EV using Level-2 charging (e.g., $6.6$kW) can lead to transformer overloading at above $115\%$ of the nominal capacity for an extra hour, potentially reducing the transformer's lifespan. Considering that EV penetration tends to cluster geographically \cite{KahnVaughn+2009}, this further exacerbates the transformer overloading problem. Hence, it becomes timely and important to understand EV charging in residential distribution systems. 

The increasing uptake of smart meters generates rich load datasets and offers promising solutions to EV charging event detection. 
Non-intrusive load monitoring (NILM) is an effective technique that has been used for load disaggregation including EV charging detection. For example, some previous studies \cite{zhang2011improved, Zhang2014} employed rule-based methods for meter-level EV charging detection. Nonetheless, these approaches suffer from limitations, including limited adaptability to diverse and dynamic load behaviors and shortcomings in terms of accuracy and generalizability. 

The rapid advances in deep learning has catalyzed the development of learning-based NILMs. These methods have made progress in addressing limitations of rule-based methods, such as improved detection accuracy \cite{meziane2017high, rehman2018low, wang2022evsense, tabatabaei2016toward, altrabalsi2016low, wang2018robust, zhou2020non}. However, the complexity and time-intensive nature of deep learning-based NILM methods pose challenges to their interpretability, restricting their potential to provide insights into EV charging behaviors.

Besides, distribution network operators may not monitor the smart meter data of individual households; instead, they collect the aggregated load of multiple households on a feeder. As a result, detecting EV charging at the feeder level emerges as a critical task of understanding and managing EV charging for the future distribution system \cite{domingos2022ev}. Recognizing the scarcity of research on EV charging detection based on aggregated load from multiple households, \emph{our work aims to bridge this gap by developing an effective algorithm for EV charging detection at the feeder level .}

This paper presents a novel framework for EV charging detection, in particular intended for distribution network operators. We synthesize feeder load data by aggregating real-world household smart meter data. Our goal is to develop a lightweight solution that relies on standard and classical machine learning algorithms, rather than focusing on complex deep learning models. We prioritize the effectiveness of input features, interpretability, and generalizability of the developed framework.
Within our developed framework, we use the Random Forest and XGBoost models for classification, i.e., charging detection. The features for the classification models are extracted based on a set of sliding windows supporting offline and online detection, respectively. The contributions of our work are summarized as follows.
\begin{itemize}
    \item We identify the challenge faced by distribution network operators in conducting feeder-level EV charging detection, distinct from household-level or meter-level EV charging detection. We address this challenge by developing a novel EV charging detection framework working on aggregated load of multiple households, achieving competitive performance.
    \item Our framework uses sliding windows to extract effective features, such as time-series statistics and the number of peaks within a time domain. Three types of sliding windows are considered, namely forward-looking, centered, and backward-looking windows. These features are demonstrated to be effective in achieving high accuracy of EV charging detection.
    \item Our framework supports both offline and online EV detection using lightweight machine learning algorithms. XGBoost achieves competitive performance in both offline and online detection, making it a well-suited algorithm for this task.
\end{itemize}
    
The rest of the paper is organized as follows. Section \ref{sub:related} reviews related works. Section~\ref{sec:problem} introduces the research problem of detecting EV charging on a feeder. Section~\ref{sec:method} presents our developed detection method, consisting of the sliding window, feature extraction, and machine learning models. Experimental results and analysis are discussed in Section~\ref{sec:results}. Section~\ref{sec:conclu} concludes our work with future research directions.

\section{Related Works}\label{sub:related}
NILM, initially introduced by Hart \cite{hart1992nonintrusive}, seeks to identify the operational states of individual appliances, distinguishing between ON and OFF states or measuring power consumption load by monitoring the aggregated electricity consumption. NILM is also considered a form of load disaggregation \cite{angelis2022nilm}. As the need for monitoring EV charging behaviors becomes increasingly pressing, more studies are using NILM for EV charging detection \cite{wang2022evsense, wang2020deep}.

Among different categories of NILM approaches for EV detection, rule-based NILM methods represent the early efforts. EV charging at the aggregate level is considered a stochastic component of the overall load. The inherent uncertainty in individual EV charging profiles presents challenges in modeling real-time EV charging behaviors across diverse scenarios \cite{qian2010modeling, li2023cross, khele2023fairness}. Therefore, domain knowledge analytics methods using rule-based NILM for EV charging detection have emerged as a potential method, e.g., \cite{zhang2011improved, Zhang2014}. For example, in \cite{Zhang2014}, a rule-based approach was presented to estimate EV charging load using a spike-train filter designed to remove power curve spikes of AC power signatures. This method does not require training and is computationally efficient. However, many rule-based methods lack validation using real-world data, and their scalability remains unclear.

In recent studies, supervised learning models have shown promising performance. A sliding window approach was often employed for extracting features from the raw active power profile \cite{meziane2017high, rehman2018low}. In \cite{8881113}, a sliding window was used to extract a mean value of the window domain comprising five samples, which was then used to evaluate the occurrence of a charging event. This work used the k-nearest neighbors (KNN) algorithm for classification, which achieved F-Scores ranging from $83\%$ to $86\%$ for EV charging events in different test scenarios. Similarly, the study in \cite{MorenoJaramillo2020} used sliding windows for feature extraction and then applied principal component analysis to mitigate noise followed by the application of random forest for classification, achieving an F-score of $92.69\%$.

In addition to EV charging detection, real-time window-based approaches have been developed for detecting appliance usage at home, harnessing the capabilities of deep learning models. In \cite{zhang2017sequencetopoint}, a sequence-to-sequence neural network was proposed for multi-class disaggregation, targeting five appliances, including dishwasher, fridge, kettle, microwave, and washing machine. Load data in sliding windows, each containing $599$ samples, were used as inputs to the model. In a similar work, \cite{NNWindows} applied neural network-based models to detect the same set of appliances, with a window length of $50$ samples at six-second intervals. The work in \cite{NNWindows} achieved F-scores of $74\%$ and $93\%$ for the fridge and kettle detection, respectively, but was unable to detect microwaves effectively. Additionally, in \cite{9310683}, a convolutional neural network was used to extract features as the inputs for a gated recurrent unit neural network (GRU). The method used a backward-looking sliding window as inputs to the convolutional layer, thereby enabling real-time detection.

Nevertheless, most prior studies have primarily concentrated on household-level NILM or appliance detection, including EV charging detection. As we explained in Section~\ref{intro}, feeder-level EV charging detection emerges as a critical task for distribution network operators but remains an unexplored area of research. 

Also, the majority of the existing studies, except for \cite{9310683}, focused on offline detection tasks as post-hoc analysis. However, certain practical scenarios may demand online or real-time detection, which can impose constraints on sliding windows. Our work is different from existing studies by developing feeder-level EV charging detection for both offline and online detection modes, achieving competitive performance compared to established household-level NILM techniques.

\section{Problem Definition} \label{sec:problem}
\begin{figure}[!ht]
    \centering
    \includegraphics[width=.99\linewidth]{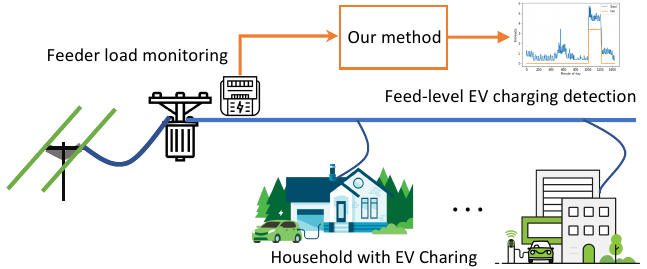}
    \caption{Illustration of non-intrusive feeder-level EV detection.}
    \label{fig:feeder_diag}
\end{figure}

This section presents the problem of feeder-level EV charging detection and delves into the intricacies and complexities inherent in EV charging detection. As depicted in Fig.~\ref{fig:feeder_diag}, the distribution network operator monitors the feeder load instead of each individual household's load to detect EV charging activities. This scenario is realistic where distribution network operators manage the distribution infrastructure, including feeders and transformers, but do not serve as the retailer supplying electricity to individual households. 

Different from household-level EV charging detection, to the best of our knowledge, feeder-level EV charging detection is not well-studied in the literature. Furthermore, feeder-level EV charging detection is more challenging, because it involves analyzing aggregated signals from multiple households, which contain a rich array of appliance usage information, as shown below.
\begin{enumerate}
    \item \textit{Aggregated Data}: Feeder load refers to the aggregated electricity consumption on one feeder hosting multiple households. The number of households varies depending on the type of distribution networks and their locations.
    The aggregated feeder load contains more noises and combinations of appliance electricity consumption across households, making EV charging detection within this context a more challenging task.
    
    \item \textit{Heterogeneous EV charging behaviors}: EV charging behaviors present another challenge because EV charging varies depending on the vehicle's state of charge and user charging behaviors \cite{Quiros-Tortos2016}. The mix of different EV charging behaviors within a single data series adds higher complexity to the detection task.
\end{enumerate}

We represent consecutive aggregated electricity consumption records over a specific time duration with the following vector: 
\begin{equation}
    \mathbf{x} = [x_1, ..., x_t, ... , x_T],
\end{equation}
where $x_t \in \mathbb{R}^+$ for $1 \leq t \leq T$, denoting the feeder-level electricity consumption at time step $t$. Similarly, we denote the corresponding EV charging electricity consumption within the same time frame as the vector:
\begin{equation}
    \mathbf{e} = [e_1, ..., e_t, ... , e_T].
\end{equation}

Note that $\mathbf{e}$ is hidden within $\mathbf{x}$ and is not directly available. We aim to use $\mathbf{x}$ to identify the EV charging state $\mathbf{y}$ over the same time span, shown as
\begin{equation}
    \mathbf{y} = [y_1, ..., y_t, ... , y_T],
\end{equation}
where $y_t$ is binary at time step $t$, i.e., $y_t \in \{0, 1\}, 1 \leq t \leq T $. When there is an EV charging event, it is ON (i.e., $y_t = 1$); otherwise, it is OFF (i.e., $y_t = 0$).

Applying supervised learning, the EV detection model $\Phi$ is trained by the training data $\mathbf{x}_{train}$ and corresponding EV charging event labels $\mathbf{y}_{train}$. Given a feeder-level load sequence $\hat{\mathbf{x}}$, the model's prediction of active EV charging state at target time step $\hat{t}$ is represented as $\Phi(\hat{\mathbf{x}})$.

Our work studies both offline and online cases of EV charging detection. In the offline scenario, the data in the past, present, and future of the target time step $\hat{t}$, shown as 
\begin{equation}
    \hat{\mathbf{x}}_{\text{offline}}= [\hat{x}_1, ..., \hat{x}_{\hat{t}},...\hat{x}_T],   1 < \hat{t} < T,
\end{equation}
can be utilized to make detection, which is a common approach adopted in many existing studies \cite{zhang2011improved, Zhang2014, cao2021smart}. 

In contrast, the online scenario restricts the use of data to only past and present observations relative to $\hat{t}$, shown as
\begin{equation}
    \hat{\mathbf{x}}_{\text{online}}= [\hat{x}_1, ..., \hat{x}_{\hat{t}}], 
\end{equation}
which can be used as input to the classifier. This limitation prevents future information (relative to $\hat{t}$) from being used for detection, which suits the need of real-time detection tasks but is yet more challenging.

\section{Methodology} \label{sec:method}
In this section, we introduce a novel EV charging detection method that places a particular emphasis on feature extraction. We develop a sliding window framework for feature extraction, transforming the time-series load data from multiple households into a feature set. To effectively identify EV charging patterns, we consider various features extracted from different types of windows, namely forward-looking, centered, and backward-looking sliding windows. These features range from basic statistics to peaks and events identified using a peak-counting algorithm. We explore these three window types, and we combine the features from three windows in certain ways to suit different scenario requirements, e.g., offline and online detection. In order to harness interpretable and lightweight machine learning models, we employ two classical machine learning models, specifically Random Forest and XGBoost, as classifiers to further validate the effectiveness, generalizability, and adaptability of the proposed method.

\subsection{Sliding Window-Based Feature Extraction} \label{slidingwindow}
Feeder-level load data provides limited data abundance, with each record containing a time step and its corresponding aggregated electricity consumption. The primary research question revolves around the extraction of features that can effectively capture EV charging patterns and facilitate the implementation of an efficient, lightweight, and interpretable method.

In this subsection, we present our designed non-intrusive load monitoring method for EV charging detection, which incorporates innovative sliding windows with different combinations tailored to offline and online detection scenarios. Furthermore, our approach develops an efficient feature extraction process. Different from existing sliding window approaches, e.g., in \cite{9310683,krystalakos2018sliding}, our method utilizes the time step of the target data points as the center and employs an adaptive window length. We create three distinct types of windows that correspond to preceding events and subsequent events relative to the target time, thus capturing rich information to suit offline and online detection scenarios.

\subsubsection{Sliding Windows}
\begin{figure}[!t]
    \centering
    \includegraphics[width=.8\linewidth]{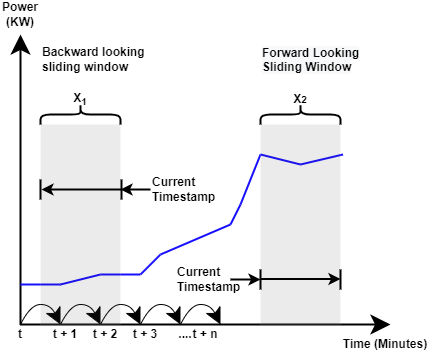}
    \caption{Illustration of forward- and backward- looking sliding windows.}
    \label{fig:window}
\end{figure}

To construct an effective feature set, we propose a sliding window-based feature extraction approach. An illustration of the sliding window is presented in Fig.~\ref{fig:window} for both the backward and forward configurations. Each sliding window covering certain length of time-series load moves along with the time step $t$. A set of features $\boldsymbol{\chi}$ is extracted based on a predefined sliding window size $n$, which is a hyper-parameter of our method. The feature set $\boldsymbol{\chi}$ from each sliding window contains the aggregated values of mean, variance, standard deviation, minimum, maximum, and median values of the time-series load, which form the basis of the feature set. Note that peak counting is also included in the feature set, and we will present it later in this section. 

We consider three distinct sliding windows as follows.
\begin{enumerate}
    \item \emph{The forward-looking sliding window.} The forward-looking sliding window initiates at time step $t$ and extends its right edge to $t + n$, in which the window size $n$ is tunable. This window type analyzes the active power profile from the present into the future relative to $t$, making it suitable for offline detection when all the data are available.
    
    \item \emph{The centered sliding window.} This case assesses both future data and past data. Specifically, the window size is halved and rounded up, denoted as $\ceil{size/2}$. If an offset parameter is provided, this offset is used to define the left edge and right edge of the window. 
    The offset shifts the sliding window to the left or right, enabling the capture of more data points from either the past or future, depending on specified needs.
    
    \item \emph{The backward-looking sliding window.} This case only assesses past data relative to $t$ with its left edge at $(t - n)$. 
As this window type only involves past data, it suits online detection tasks very well.
\end{enumerate}
\bigbreak

\subsubsection{Peak Counting}
Our work also uses a sliding window-based peak-counting algorithm to identify the number of peaks as a feature. This algorithm analyzes the time-series load data and sets a single parameter: a threshold $\theta$. For instance, a centered sliding window is employed with a window size of $360$. Within the given window, the peak counting algorithm calculates the count of peaks. Specifically, given a sequence of input $\mathbf{x}$, the total number of peaks is determined by
\begin{equation}\label{peak_eq}
    P(\mathbf{x}) = \sum^N_{t=1}[\frac{x_t - x_{t-1}}{x_{t-1}}> \theta],
\end{equation} 
where $x_t$ represents the current value within the input sliding window and $x_{t-1}$ is the previous value. As for the threshold $\theta$, cross-validation is employed to find an appropriate value, and it is set to $1$ in this paper.

\subsection{Features for Online and Offline Cases}
In the offline scenario for feeder-level detection, all three types of sliding windows are utilized. Within these windows, we extract features, including the mean, standard deviation, variance, minimum, maximum, median, and the count of peaks. 
\begin{enumerate}
    \item The first window is a centered window with half of the window domain containing samples from the past and future relative to the current time step.
    \item The second window is a forward-looking sliding window, where the samples extend from the current time step into the future.
    \item The third window is a backward-looking sliding window. In contrast, the samples are from the past up to the current time step.
\end{enumerate}
The window length is set at $360$ based on the hyper-parameter optimization in our numerical analysis, as detailed in Section \ref{sec:results}.

In the online framework, we employ two passes of the backward-looking sliding windows with two different lengths. Specifically, the first window has a length of $360$, and the second window has a length of $90$. Additionally, the online case incorporates the count of peaks based on a backward-looking window with a length of $360$.
The reason for using multiple windows is that the window length has a direct impact on performance depending on the duration of the appliance load \cite{10.1145/2821650.2821672}. EV charging profiles can vary in length due to factors such as the vehicle’s state of charge and charging behavior. Some charging events can be long-lasting, lasting for several hours, while others can be shorter. Therefore, employing multiple windows of varying lengths can enhance performance by better capturing charging events of shorter duration.

\subsection{Machine Learning Models}
To ensure a balance between robust performance and interpretability, our approach employs classical machine learning models for classification tasks. The sliding window algorithm generates a set of features, which are subsequently employed as inputs for these classification models. Specifically, XGBoost serves as the primary classification model, with Random Forest acting as a baseline for comparison. We choose XGBoost \cite{10.1145/2939672.2939785} due to its swift training capabilities and its well-established reputation for delivering top-tier performance in structured data problems. Importantly, it is worth noting that our aim is not to develop novel machine learning models but to employ classical models to evaluate the effectiveness of our sliding-window approach and feature extraction.

\begin{table}
\centering
\large
\caption{Data description}
\label{data_sta}
\resizebox{0.45\textwidth}{!}{\begin{tabular}{lcc}
\hline
 &Individual Household Meter &  3-household Feeder   \\
\hline
Number of measurement items & 85 &  22  \\ 
Number of records & 32,795,200 &  11,072,891 \\
Number of EV charging events & 15,654  & 12,728 \\
\hline
\end{tabular}}
\end{table}

\section{Experimental Evaluation}\label{sec:results}

\subsection{Experimental Settings}
We use minute-interval residential load data with EV charging labels 
over one year from Pecan Street \cite{street2016pecan}. As there was an absence of labeled feeder-level data, we generate synthetic feeder-level load data by randomly aggregating the load data from several households. \footnote{The number of households for synthesizing feeder-level load depends on the type of network and local practices. We start with $3$ households to simulate the Australian scenario in this work, while we plan to explore different scenarios in our future research.} The feeder-level EV charging labels are derived from the household meter data, where an EV charging label is positive if one or more households have EV charging event(s) at a given time step. As shown in Table \ref{data_sta}, meter data of $85$ households with more than thirty million records are used, and $22$ feeders' synthetic load are generated. In our experiment, we train the model using synthetic records from some feeders to detect EV events in other feeders. We partition the data into 75\% for training, 15\% for validation, and 10\% for testing, based on the feeder number.

\subsection{Performance Metrics for EV Charging Detection}
We assess model performance using standard classification metrics, namely accuracy, precision, recall, and f-score. Since the EV charging events are spare, accuracy alone might be misleading, as it could be always high even actual EV charging events are not detected by classifiers. To better capture the detection performance, we focus on the following three metrics:
\begin{align}
    & \text{Recall} = \dfrac{TP}{TP+FN}, \label{recall} \\
    & \text{Precision} = \dfrac{TP}{TP+FP}, \label{Precision} \\
    & \text{F}_{1}= 2 \times \frac{\text{Precision} \times \text{Recall}}{\text{Precision} + \text{Recall}}, \label{f1}
\end{align} 
where $TP$, $FP$, and $FN$, are true positive, false positive, and false negative, respectively. Note that precision and recall are more suitable metrics for our problem as they provide insights into the detection of charging events. The $F_1$ score is the balanced combination of precision and recall and is helpful in evaluating model performance as a single metric, especially for unbalanced data.

\begin{figure}
    \centering
    \includegraphics[width=.8\linewidth]{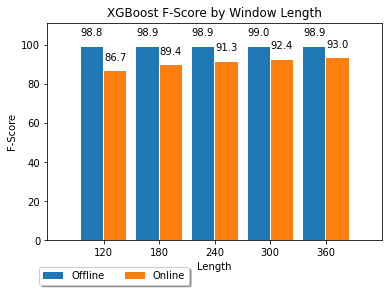}
    \includegraphics[width=.8\linewidth]{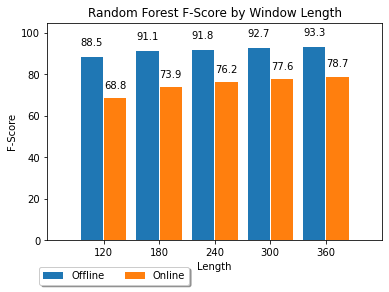}
    \caption{F-Score for XGBoost and random forest by window lengths.}
    \label{fig:f1_by_length}
\end{figure}

\subsection{Results and Discussions}
In this subsection, we present experimental results and findings, providing valuable insights into the performance and robustness of our method.

\subsubsection{Window Length Affect Studies}
The size of the sliding windows significantly impacts the detection performance when using sliding window-based feature extraction methods.
The results shown in Figure \ref{fig:f1_by_length} indicate that the selection of window length influences the performance of EV charging detection. Longer sliding windows lead to improved performance on both online and offline detection using XGBoost and random forest. However, it's noteworthy that the improvement becomes marginal as the window length approaches $360$, i.e., $6$ hours. Moreover, there is a trade-off between the window size and the computational resources required, as larger windows include more data and demand more processing time. This trade-off should be carefully considered when determining the window size for EV charging detection. In our work, we have chosen a window length of $360$.

\subsubsection{Models Comparisons} 
The performance comparison of XGBoost and random forest models are presented in Table~\ref{tab:performance}. Both models perform well in the offline setting, achieving $F_1$ Scores of 98.88\% and 93.28\%, respectively, with similar precision and recall results. XGBoost outperforms random forest in terms of recall and $F_1$, albeit with a slightly recall. 

For the online detection, where only historical data up to the current time step can be used, XGBoost maintains an $F_1$ Score of 93.01\%. In contrast, random forest achieves an $F_1$ Score of $78.7\%$. Despite the limited features extracted in the online scenario, XGBoost continues to achieve competitive performance compared to the offline scenario.

\begin{table}
\centering
\small
\caption{Performance comparison of XGBoost and random forest in both offline and online scenarios.}
\label{tab:performance}

\begin{tabular}{|l|l|l|l|}
\hline
 \textbf{\begin{tabular}[c]{@{}l@{}}Model \end{tabular}} & \textbf{\begin{tabular}[c]{@{}l@{}}Metrics \end{tabular}}  &
  \textbf{\begin{tabular}[c]{@{}l@{}}Offline \end{tabular}} &
  \textbf{\begin{tabular}[c]{@{}l@{}}Online \end{tabular}} \\ 
  \hline
  \multirow{3}{4em}{XGBoost} & Precision     & 98.96\% & 94.39\% \\
  & Recall        & \textbf{98.80\%} & \textbf{91.66\%} \\
  & $F_1$         & \textbf{98.88\%} & \textbf{93.01\%} \\ 
\hline
 \multirow{3}{4em}{Random Forest} & Precision     & \textbf{99.47\%} & \textbf{96.51\%} \\
 & Recall        & 87.83\% & 66.43\% \\ 
& $F_1$         & 93.28\% & 78.70\% \\ \hline
\end{tabular}
\end{table}

\section{Conclusion and Future Work} \label{sec:conclu}
This paper presented a sliding window-based NILM method for feeder-level EV charging detection. By analyzing synthetic feeder-level load data, our method extracts effective features, including statistical features and peak counts. We employ classical classification models to detect EV charging events, making use of sliding window-based features. Furthermore, this paper investigated the applicability of these feature engineering techniques to online detection, achieving competitive performance compared to offline detection. The numerical results demonstrated that our method offers a practical solution for EV charging detection on feeders, providing valuable insights for the distribution system operator.

In future work, we plan to delve deeper into the NILM for the feeder-level detection with a broader range of households and examine its effectiveness.

\bibliographystyle{unsrt}
\bibliography{ref.bib}

\end{document}